%% file: main.tex
\documentclass{article}

\PassOptionsToPackage{numbers, compress}{natbib}
\usepackage[preprint]{neurips_2022}

\usepackage[utf8]{inputenc} % allow utf-8 input
\usepackage[T1]{fontenc}    % use 8-bit T1 fonts
\usepackage{hyperref}       % hyperlinks
\usepackage{url}            % simple URL typesetting
\usepackage{booktabs}       % professional-quality tables
\usepackage{amsfonts}       % blackboard math symbols
\usepackage{nicefrac}       % compact symbols for 1/2, etc.
\usepackage{microtype}      % microtypography
\usepackage{xcolor}         % colors

\usepackage{wrapfig}
\usepackage{algorithm}
\usepackage{algorithmic}
\usepackage{float}
\newcommand{\spub}{s_{\tiny{\mbox{pub}}}}
\newcommand{\Spub}{\mathcal{S}_{\tiny{\mbox{pub}}}}

\definecolor{green}{HTML}{4DAF4A}

\newcommand*\samethanks[1][\value{footnote}]{\footnotemark[#1]}

\usepackage{graphicx}
\usepackage{enumitem}
\usepackage{wrapfig}
\usepackage{subcaption}
\usepackage{amsthm}

\usepackage{thmtools}

\declaretheoremstyle[
  bodyfont=\normalfont\itshape,
  headformat=\NAME\NUMBER  
]{nospacetheorem}
\declaretheorem[style=nospacetheorem,name=Assumption]{assumption}
\declaretheorem[style=nospacetheorem,name=Definition]{definition}

\newcommand{\ourmethod}{IMPROVISED}
\usepackage{mathtools}

\usepackage{amsmath}
\DeclareMathOperator{\argmax}{argmax}

\DeclareMathOperator{\softmax}{softmax}

\title{Self-Explaining Deviations for Coordination}

\author{
Hengyuan Hu \thanks{Equal Contribution.} \\ Meta AI \And
Samuel Sokota \samethanks \\ Carnegie Mellon University  \And
David Wu \\ Meta AI  \And
Anton Bakhtin\\ Meta AI \And
Andrei Lupu \\ McGill University \And
Brandon Cui \\ Meta AI \And
Jakob N. Foerster \thanks{Work done while at Meta AI.}\\ University of Oxford }

\begin{document}

\maketitle

\input{0_abstract}
\input{1_intro}

\input{2_background}
\input{3_sed}

\input{4_improvised}
\input{5_experiment}
\input{6_related_work}

\input{7_conclusion}
\newpage

\bibliographystyle{abbrvnat}
\bibliography{neurips_2021.bib}

\newpage
\appendix
\input{A_algo}

\end{document}

%% file: 0_abstract.tex
\begin{abstract}
Fully cooperative, partially observable multi-agent problems are ubiquitous in the real world.
In this paper, we focus on a specific subclass of coordination problems in which humans are able to discover \emph{self-explaining deviations} (SEDs).
SEDs are actions that deviate from the common understanding of what reasonable behavior would be in normal circumstances.
They are taken with the intention of causing another agent or other agents to realize, using theory of mind, that the circumstance must be abnormal.
We first motivate SED with a real world example and formalize its definition. 
Next, we introduce a novel algorithm, improvement maximizing self-explaining deviations (IMPROVISED), to perform SEDs. 
Lastly, we evaluate IMPROVISED both in an illustrative toy setting and the popular benchmark setting Hanabi, where it is the first method to produce so called \emph{finesse} plays, which are regarded as one of the more iconic examples of human theory of mind.

% Fully cooperative, partially observable multi-agent problems are ubiquitous in the real world.
% These problems may require agents to collaborate with agents that were trained separately, i.e., to perform zero-shot coordination.
% In this paper, we focus on a specific subclass of zero-shot coordination (ZSC) problems in which humans are able to discover \emph{self-explaining deviations} (SEDs).
% SEDs are actions that deviate from the common understanding of what reasonable behavior would be in normal circumstances.
% They are taken with the intention of causing another agent or other agents to realize, using theory of mind, that the circumstance must be abnormal.
% First, we extend the ZSC setting to accommodate SEDs to facilitate our investigation.
% Next, we introduce a novel algorithm,  IMPROVement maxImizing Self-Explaining Deviations (IMPROVISED), designed to perform SEDs. 
% Lastly, we evaluate the efficacy of IMPROVISED both in an illustrative toy setting and the popular benchmark setting Hanabi, where it is the first learning method to produce so called \emph{finesse} plays, widely regarded as one of the more iconic examples of human theory of mind.
\end{abstract}

%% file: 1_intro.tex
\section{Introduction}

\begin{wrapfigure}{r}{0.45\textwidth}
\centering
% \begin{minipage}{.5\textwidth}
\centering
\renewenvironment{quote}
{\small\list{}{\rightmargin=0.25cm \leftmargin=0.0cm}%
\item\relax}
{\endlist}
\begin{quote}
\small
Dispatcher: Oregon 911.\\
Caller: I would like to order a pizza at...\\
Dispatcher: You called 911 to order a pizza?\\
Caller: Uh, Yeah, apartment...\\
Dispatcher: This is the wrong number to call for a pizza.\\
Caller: No no no... you're not understanding me.\\
Dispatcher: I'm getting you now. Is the other guy still there?\\
Caller: Yep. I need a large pizza.\\
Dispatcher: All right. How about medical. You need medical?\\
Caller: No. With pepperoni.\\
Dispatcher: Turn your sirens off before you get there. Caller ordered a pizza. And agreed with everything I said that there's domestic violence going on.
\end{quote}
\caption{{\small A real-life self-explaining deviation.}}
\label{fig:pizza}
\end{wrapfigure}

There has been a lot of progress in the deep multi-agent collaboration domain ranging from better learning~\cite{yu2021surprising, sad} and planning~\cite{sparta,sokota2021solving} algorithms designed to achieve strong performance in self-play to finding policies that collaborate well with other agents include humans~\cite{otherplay, fict-coplay}. However, these algorithms largely assume that the partners at test time will follow a fixed policy. Even in one of the most powerful search algorithms~\cite{sparta}, the agent performing search assumes that other agents follow blueprint when planning forward, resulting in primarily unilateral deviations that improve the performance without relying on other players also deviating in response.

Humans generally assume other human follow certain social norms (blueprint policies) when acting in the society, and interpret other humans' behavior based on such norms to infer hidden information about the partially observable world. 
For example, when a driver sees a car ahead suddenly stopping on a road, they may infer that there is some accident ahead or the car is broken. 
However, human also have the ability to improvise when the conventional actions are restricted or there exist other actions that can bring superior outcomes in situations. 
Taking the event in Figure~\ref{fig:pizza} as an example, the caller deviates from the conventional practice of stating the situation truthfully because it has negative consequences. 
In this situation, the caller would perform this deviation because the deviation is only beneficial if the dispatcher figures out caller's real intention and also deviates instead of simply ending the call.
This type of deviation is interesting because, firstly, it can be detected by the other person because it appears to be a mistake under common practice and, secondly, the expected deviation for the other person in response can be independently decided by both individuals based on the common understanding of the world and the deviation chosen by the first person. We refer to this type of phenomena as self-explaining deviations (SEDs).

SEDs are actions that, under normal circumstances, would not make sense, given the agents' common understanding.
They are executed with the intention that teammates will use \emph{theory of mind} to deduce that the situation is unusual in a particular and important way, and will adapt their behavior to account for this additional information.
While there is ample evidence that humans perform SEDs, such as the \emph{finesse play} in the card game Hanabi, we provide some evidence that neither existing planning nor learning algorithms can do so.

% IMPROVISE
In this paper, we first formalize the problem setting and definition of SEDs. Then we introduce a novel planning algorithm, IMPROVement maxImizing Self-Explaining Deviations (IMPROVISED), for performing SEDs.
We show that, under some assumptions, IMPROVISED performs the optimal SED in terms of expected return maximization.
Next, we provide a motivating experiment, illustrating that in a small toy problem designed to require SEDs to perform optimally, IMPROVISED is able to compute an optimal joint policy, whereas other multi-agent algorithms are not.
Lastly, we present experiments on Hanabi, where we show that IMPROVISED is able to produce finesse plays, which is one of the most interesting techniques that human experts perform frequently.

%% file: 2_background.tex
\section{Background}

\textbf{FOSG and Public POMDP.} For our notation, we use an adaption of factored observation stochastic games (\textbf{FOSG}) \citep{fosg}.
$\mathcal{W}$ is the set of \textbf{world states} and $w^0$ is a designated initial state. 
$\mathcal{A}= \mathcal{A}_1 \times \cdots \times \mathcal{A}_N$ is the space of \textbf{joint actions}.
$\mathcal{T}$ is the \textbf{transition function} mapping $\mathcal{W} \times \mathcal{A} \to \Delta(\mathcal{W})$. 
$\mathcal{R} \colon \mathcal{W} \times \mathcal{A} \to \mathbb{R}$ is the \textbf{reward function}. 
$\mathcal{O} = (\mathcal{O}_{\mbox{\tiny priv}(1)}, \dots, \mathcal{O}_{\mbox{\tiny priv}(N)}, \mathcal{O}_{\mbox{\tiny pub}})$ is the \textbf{observation function} where $\mathcal{O}_{\mbox{\tiny priv}(i)} \colon \mathcal{W} \times \mathcal{A} \times \mathcal{W} \to \mathbb{O}_{\mbox{\tiny priv}(i)}$ specifies the \textbf{private observation} that player $i$ receives. 
$\mathcal{O}_{\mbox{\tiny pub}} \colon \mathcal{W} \times \mathcal{A} \times \mathcal{W} \to \mathbb{O}_{\mbox{\tiny pub}}$ specifies the \textbf{public observation} that all players receive.
$O_i{=}\mathcal{O}_i(w, a, w'){=}(\mathcal{O}_{\mbox{\tiny priv}(i)}(w, a, w'), \mathcal{O}_{\mbox{\tiny pub}}(w, a, w'))$ is player $i$'s \textbf{observation} and a \textbf{history} is a finite sequence $h=(w^0, a^0, \dots, w^t)$.
The \textbf{set of histories} is denoted by $\mathcal{H}$. 
The \textbf{information state} for player $i$ at $h = (w^0, a^0, \dots, w^t)$ is $s_i(h) \coloneqq (O_i^0, a_i^0, \dots, O_i^t)$. 
The \textbf{information state space} for player $i$ is $\mathcal{S}_i \coloneqq \{s_i(h) \mid h \in \mathcal{H}\}$. 
The \textbf{legal actions} for player $i$ at $s_i$ is denoted $\mathcal{A}_i(s_i)$. 
A \textbf{joint policy} is a tuple $\pi = (\pi_1, \dots, \pi_N)$, where \textbf{policy} $\pi_i$ maps $\mathcal{S}_i \to \Delta(\mathcal{A}_i)$. 
The \textbf{public state} at $h$ is the sequence $\spub(h) \coloneqq \spub(s_i(h)) \coloneqq (O_{\mbox{\tiny pub}}^0, \dots, O_{\mbox{\tiny pub}}^t)$. 
% The \textbf{public tree} $\Spub \coloneqq \{\spub(h) \mid h \in \mathcal{H}\}$ is the space of public states.
The \textbf{information state set} for player $i$ at $s \in \Spub$ is $\mathcal{S}_i(s) \coloneqq \{s_i \in \mathcal{S}_i \mid \spub(s_i) = s\}$, where $\Spub$ is the space of public states. 
Finally, the \textbf{reach probability} of $h$ under $\pi$ is $P^{\pi}(h)$.

Rather than working with in a multi-agent setting directly, we invoke the public POMDP transformation that maps cooperative multi-agent settings to equivalent single-agent POMDPs.
Given a common-payoff FOSG $\langle \mathcal{N}, \mathcal{W}, w^0, \mathcal{A}, \mathcal{T}, \mathcal{R}, \mathcal{O} \rangle$, we can construct an equivalent \textbf{public POMDP}~\citep{nayyar} $\langle \tilde{\mathcal{W}}, \tilde{w}^0, \tilde{\mathcal{A}}, \tilde{\mathcal{T}}, \tilde{\mathcal{R}}, \tilde{\mathcal{O}} \rangle$ as follows: 
The world states of the public POMDP $\tilde{\mathcal{W}}$ is the set
$\{\left(s_1(h), \dots, s_N(h)\right) : h \in \mathcal{H}\}$. The initial world state of the public POMDP $\tilde{w}^0$ is the tuple 
$(s_1(h^0), \dots, s_N(h^0))$. The actions of the public POMDP are called \textbf{joint decision rules}. It is denoted by $\Gamma$ and has $N$ components.
The $i$th component of it $\Gamma_i$ is the \textbf{decision rule} for player $i$. 

A decision rule $\Gamma_i$ maps $s_i$ to an element of $\mathcal{A}_i(s_i)$ for each $s_i \in \mathcal{S}_i(\spub(h))$;
it instructs a player in the common-payoff FOSG how to act as a function of its private information. Given $\tilde{w} \equiv (s_1, \dots, s_n)$ and $\Gamma$, the transition distribution $\tilde{\mathcal{T}}(\tilde{w}, \Gamma)$ is induced by $\mathcal{T}((s_1, \dots, s_n), a)$, where $a \equiv \Gamma((s_1, \dots, s_n)) \coloneqq \left(\Gamma_1(s_1), \dots, \Gamma_N(s_N)\right)$. Given $\tilde{w} \equiv (s_1, \dots, s_n)$ and $\tilde{w}'\equiv (s_1', \dots, s_n')$, 
the reward and observation are given by $\tilde{\mathcal{R}}(\tilde{w}, \Gamma, \tilde{w}') \equiv \mathcal{R}((s_1, \dots, s_n), \Gamma((s_1, \dots, s_n)), (s_1', \dots, s_n'))$
and $\tilde{\mathcal{O}}(\tilde{w}, \Gamma, \tilde{w}') \equiv \mathcal{O}_{\mbox{\tiny pub}}((s_1, \dots, s_n), \Gamma((s_1, \dots, s_n)), (s_1', \dots, s_n'))$,
respectively.

Every policy in the public POMDP corresponds to a joint policy in the underlying common-payoff FOSG, which receives exactly the same expected return.
Therefore, it is sufficient to work with the public POMDP.
See \citep{sokota_solving_2020,sokota2021solving} for further discussion.

When the public POMDP is considered as a belief MDP, rather than as a POMDP, its belief states are of the form $b^t \colon (s_1^t, \dots, s_n^t) \mapsto P^{\Gamma^0, \dots, \Gamma^{t-1}}(s_1^t, \dots, s_N^t \mid \spub^t)$.
In words, this is the joint distribution over private information states, conditioned on the historical policy $(\Gamma^0, \dots, \Gamma^{t-1})$ and the public state $\spub^t$.
This public belief MDP is abbreviated as the PuB-MDP.

% \textbf{Zero-Shot Coordination.} Procedurally, the zero-shot coordination setting \citep{otherplay} investigates the performance of agents trained independently and evaluated via cross-play.
% A ``high-level description'' of a learning algorithm $\mathcal{L}$ is provided to multiple experimenters $\mathcal{E}_1, \dots, \mathcal{E}_N$.
% Each experimenter $\mathcal{E}_i$ implements its own version of algorithm $\mathcal{L}$, and trains a team of agents $\alpha_{i,1}, \dots, \alpha_{i, N}$ using its implementation of $\mathcal{L}$ in self-play.
% The evaluation of $\mathcal{L}$ is determined by the performance of the team $\alpha_{j_1, 1}, \dots, \alpha_{j_N, N}$ where $j_1, \dots, j_N$ is a random permutation of $1, \dots, N$.
% As a matter of practice, this evaluation scheme is difficult to enforce, so it is acceptable for an experimenter to train each team $\{\alpha_{i, j} : j =1, \dots, N \}$ using the same code, so long as it does not exploit hyperparameters that would have a large effect on the cross-play score, such as the seed or tie-breaking procedures.

% While a high cross-play score as defined above does not suggest that an agent would perform well with arbitrary teammates, it is indicative of ability to perform well with ``reasonable'' teammates.
% This is because requiring good cross-play prohibits algorithms using arbitrary and symmetry-breaking convention systems, which would be impossible to decode in a zero-shot setting.

%% file: 3_sed.tex
\section{Self-Explaining Deviations}
\label{sec:sedsbackground}

While the name \textit{self-explaining deviation} (SED) is novel to this work, the idea behind SEDs is not. Perhaps the most famous example of SEDs comes from the cooperative card game Hanabi \citep{hanabi} in the form of a play called a ``finesse''.
Readers unfamiliar with Hanabi and finesse may first jump Section~\ref{sec:hanabi-def} and Section~\ref{sec:whatisfinesse} for the detailed descriptions.
In a finesse play, the acting player, i.e., the first player who initiates the deviation as part of the finesse, intentionally communicates misleading information to a receiving player, which would hurt the team's score if the receiving player acted upon it using the prior convention. 
The second player in the finesse, who acts after the first player but before the receiving player, realizes that the first player must have deviated from the prior convention after observing the seemingly disastrous move. 
% would only have misled the receiving player if it were possible for the observing player to intervene such that the team instead benefits.
However, knowing that the first player is rational, the second player realizes that they can also deviate to reach a better outcome than the one that the original convention would have led to. After the second player plays its part of the joint deviation, the original information from the first player is no longer misleading and the subsequent players can continue to follow the prior conventions.

SEDs may take place in any cooperative situation with at least two players where common knowledge blueprint policies that players follow under normal assumptions exist. 
% Figure~\ref{fig:pizza} shows a real world example. 
Intuitively, SEDs capture a form of joint deviations where one agent takes an action that at first appears to be a mistake or otherwise highly off-policy from another player's perspective, i.e. zero or low probability under the blueprint. On the presumption that the first agent chose that action intelligently and deliberately and there is a unique off-policy action for the observing agent that could potentially result in an even-better-than-normal outcome for both, the observing agent may reason that the first agent ``intends'' them to take it---to take a leap of faith that the first agent has not erred, but rather knows both agents can get that better outcome, even if the observing agent does not have the information themselves to prove that this outcome will result.
% Despite the fact the Hanabi players often have never met or communicated prior to playing together, successful finesses as well as other simpler SEDs are a regular occurrence in online play, with independent discovery of finesses known to have happened in multiple player-groups.
% As a result, multi-agent learning researchers regard the finesse play as an example of humans performing theory of mind.
This work uses the term SEDs to describe this phenomenon of communication via apparent mistakes in a general context. We formalize SED in Section~\ref{sec:examiningseds} and propose IMPROVISED, a novel planning algorithm that performs the optimal SEDs under some assumptions, in Section~\ref{sec:improvised}.
% \footnote{To give some further intuition about what an SED is - consider what the example in Figure 1, finesses in Hanabi discussed further in Section \ref{sec:whatisfinesse}, and the Trampoline-Tiger game in Section \ref{sec:examiningseds} all have in common. In all of them, agent A acts in a way that appears to agent B to be a mistake. Only if B \emph{trusts} that A's action was not a mistake, then B, even in zero-shot, can identify a unique most-plausible way to help. However, A's action not being an mistake relies on B being likely to make this deduction. This seeming circularity - the need for a \emph{joint} deviation - makes this challenging for current algorithms, particularly in zero-shot settings.}

% Lastly, we note that in any given population of agents, if the SED action and response is part of the blueprint policy already, or if an SED is discovered and becomes common enough that the agents learn and incorporate it into the blueprint, it will quickly become the equilibrium and no longer be a deviation at all. However such an equilibrium may not be common among agents or easy to find otherwise (e.g. as we see in Section \ref{sec:trampolineresults}). Our work can therefore be viewed as an exploration of a novel policy improvement operator that humans can intuitively apply even in zero-shot settings that current algorithms are largely incapable of.

% \end{minipage}%
% \begin{minipage}{.5\textwidth}
\begin{wrapfigure}{r}{0.5\textwidth}
    \vspace{-4mm}
    \centering
    \includegraphics[width=.48\textwidth]{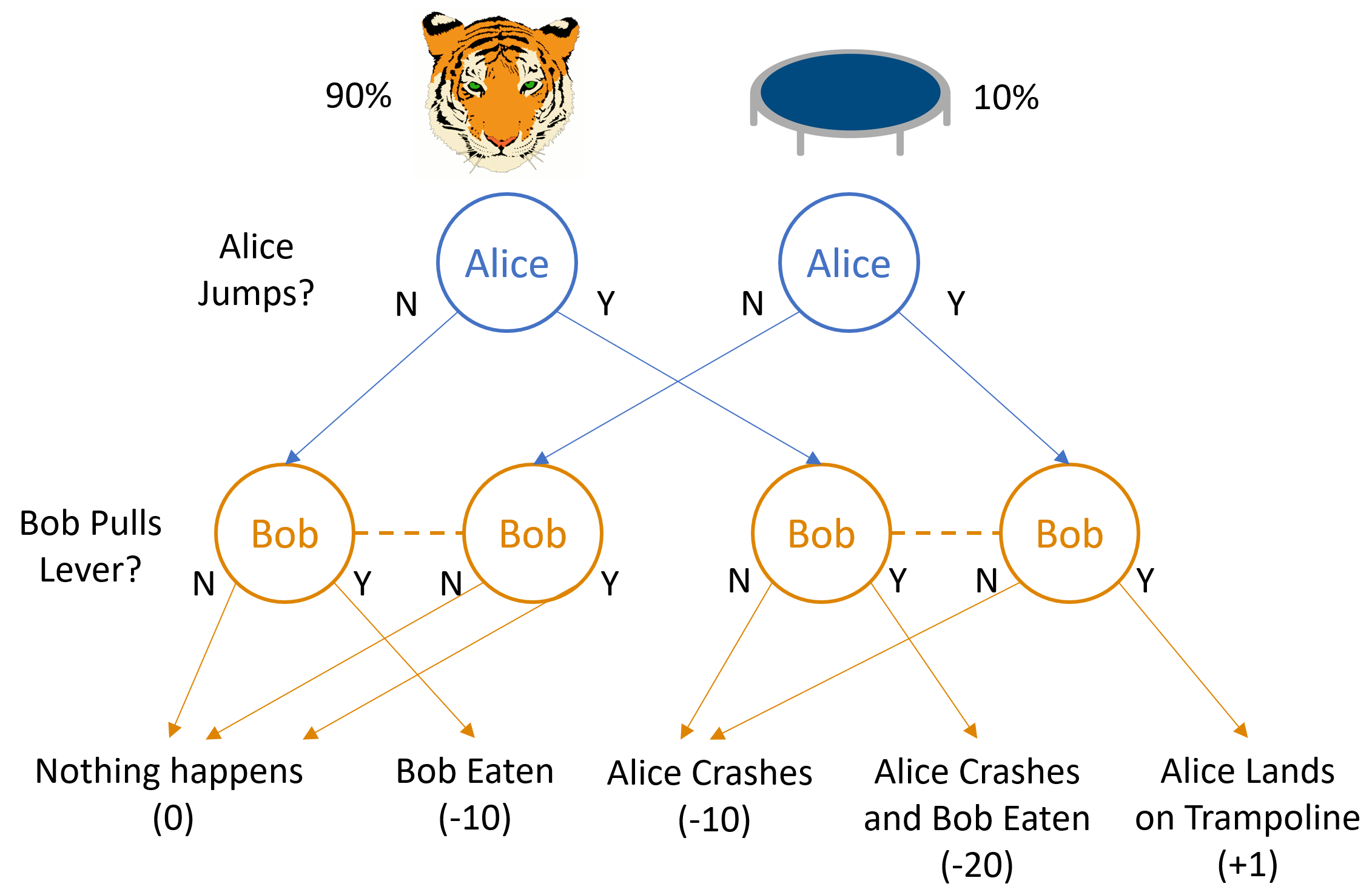}
    \caption{\small Alice knows whether Bob's lever will deploy a tiger or a trampoline, and then jumps or not. Bob observes only Alice's action and then pulls the lever or not. Dotted lines between two Bob nodes means that Bob cannot distinguish them.}
    % \vspace*{-6mm}
    \label{fig:trampoline_tiger}
    \vspace{-6mm}
    \label{fig:teaser}
\end{wrapfigure}

\subsection{Examining Self-Explaining Deviations}
\label{sec:examiningseds}
Having defined the problem setting, we are now ready to investigate SEDs.
To facilitate our investigation, let's consider the \emph{trampoline-tiger} game in Figure \ref{fig:trampoline_tiger} as an example.
In the game, Alice is standing on a balcony, while Bob stands on the ground next to a lever. Pulling the lever will either deploy a trampoline below the balcony, or release a tiger. From above, Alice can observe which is the case, and then Alice decides whether to jump off the balcony or not. Bob \emph{observes Alice's choice} but does \emph{not} know whether the lever will deploy a trampoline or tiger, and decides to pull the lever or not. Alice wants to get down from the balcony, but will die from the fall unless Bob pulls the lever \emph{and} it releases a trampoline. If Bob pulls the lever and releases a tiger, it will eat Bob.

Consider the joint policy $(\ast \mapsto \text{N}, \ast \mapsto \text{N})$.
While not optimal, this joint policy is not unreasonable, as Bob won't be eaten and Alice won't fall to her death.
Indeed, independent of the probability of a trampoline, this joint policy is a Nash equilibrium. In addition, as we shall see in Section~\ref{sec:trampolineresults}, state-of-the-art MARL algorithms converge to this solution most of the time.
However, there is a clear opportunity for a SED under this blueprint.
If there is a tiger, Alice should obviously never jump. 
So if Bob observes Alice choosing to jump, and trusts that Alice is intelligent and would only have chosen to jump if it could lead to a better result, Bob should realize that there must be a trampoline. 
Therefore Alice should trust that Bob will pull the lever if she jumps, as a result.

Under any planning algorithm that considers only unilateral deviations, Bob and Alice can't escape this local optimum. Alice will never jump because she is under the belief that Bob will choose not to pull the lever.
Therefore, finding SEDs in general requires considering multi-lateral deviations, i.e. simultaneous deviations by more than one player.

Since finding multi-lateral deviations in general games with partially observable state is a complex problem, we restrict our focus to settings satisfying the following assumptions:

\begin{assumption} \label{rmk:pub}
We assume sequential and publicly observable actions.
\end{assumption}
SEDs may still be possible with partially-observable actions, but for simplicity we focus on when Bob can directly observe Alice deviate from the blueprint, rather than having to deduce it from partial or incomplete observations.

\begin{assumption} \label{rmk:action}
We consider only cases where the Bob's private information is not necessary for coordinating on the SED.
\end{assumption}
For simplicity, we always refer to the first player who initiates the joint deviation as the {\it Alice} and the second player who figures out Alice's intention and plays their part of the joint deviation as {\it Bob} in the following discussions beyond the scope of this toy game. In other words, we only consider SEDs where Bob needs only act as a function of the common-knowledge state and the Alice's action, rather as a function of private knowledge of their own.

Under these assumptions, we define the SED as follows:
\begin{definition} \label{rmk:def}
Given a common knowledge blueprint (BP) policy $\pi$, common public belief of Alice and Bob $b$, and information state of Alice $s_1$, SED is any pair of joint deviations $(a_1', a_2')$ for Alice and Bob that satisfy all the following conditions:
\begin{itemize}
    \item $q_{\pi}(b, s_1, a_1', a_2') \geq  q_{\pi}(b, s_1)$, i.e., it gives higher expected future return than BP;
    \item $P(a_1' |~\pi, b) \leq \epsilon$, i.e., $a_1'$ is highly unlikely under $\pi$ under the common belief;
    \item $a_2' \sim f(b, a_1')$, i.e., $a_2'$ is a function of the public belief and Alice's deviation action.
\end{itemize}
\end{definition}
Here, $q_{\pi}(b, s_1, [a_1, a_2])$ is Alice's estimate of the expected future return after executing the optional $a_1, a_2$ and following $\pi$ afterwards. $\epsilon$ is a hyper-parameter. The function $f$ is determined by the algorithm designer under the constraint that $f$ only takes $b$ and $a_1'$ as input. It is used both by Alice to predict Bob's response and by Bob to decide the response independently.

%% file: 4_improvised.tex
\section{IMPROVISED}
\label{sec:improvised}

In this section we derive IMPROVISED, an algorithm to perform the SEDs. The core of the algorithm is to find a function $f$ under the constraint of SED and an optimization procedure that Alice uses to decide whether to deviate given her information state $s_1$ and possible responses $a_2 \sim f$ from Bob.

\subsection{Defining the Optimization Problem}
With the observations in the prior sections, we are ready to write selecting a SED as a PuB-MDP optimization problem:
% \begingroup
% \small
% \allowdisplaybreaks
\begin{equation}
\begin{split}
\max_{\Gamma_1, \Gamma_2} \, q_{\pi}(b, \Gamma_1, \Gamma_2) 
& = \max_{\Gamma_1, \Gamma_2} \, \mathbb{E}_{s_1 \sim b} \, q_{\pi}(b, s_1, \Gamma_1(s_1), \Gamma_2)\\
& = \max_{\Gamma_1, \Gamma_2} \, \mathbb{E}_{s_1 \sim b} \, q_{\pi}(b, s_1, \Gamma_1(s_1), \Gamma_2 \circ \Gamma_1(s_1))\\
& = \max_{\Gamma_1, \Gamma_2} \, \mathbb{E}_{s_1 \sim b} \, \max\left[q_{\pi}(b, s_1, \Gamma_1(s_1), \Gamma_2 \circ \Gamma_1(s_1)), q_{\pi}(b, s_1)\right].
\label{eq:general}
\end{split}
\end{equation}
% \endgroup
The series of equalities above begins with a generic multi-lateral search in the PuB-MDP.
The LHS is ranging over Alice and Bob's decision rules such that, if Alice uses $\Gamma_1$ and Bob uses $\Gamma_2$ in belief state $b$, and all the players play according to $\pi$ thereafter, the expected return is maximized.
The first equality holds without applying any additional assumptions, simply based on the fact that Alice is able to observe $s_1$.
The second equality holds by applying Assumptions \ref{rmk:pub} and \ref{rmk:action}, which imply both that Bob can observe Alice's action and that it is the only thing he needs to condition his decision upon (note that the change of the set over which $\Gamma_2$ ranges is left implicit).
% The final holds by applying Assumptions \ref{rmk:opt} and \ref{rmk:supp}.
The final holds due to the design decisions that a) our algorithm only plays an SED for actions not supported in the BP and that b) players opt into the BP when there is no better deviation.
The inner maximization expresses the fact that Alice can observe $s_1$ and may opt into the blueprint if the blueprint value exceeds that of the deviation.
Because Bob knows which actions are supported by the blueprint, he can opt in exactly when Alice opts in, and play according to the deviation otherwise.

\subsection{An Easier Special Case} 
\label{sec:easier}
Unfortunately, expression (\ref{eq:general}) remains difficult to optimize, as the number of $\Gamma_1$ and $\Gamma_2$ is combinatorial.
% To ameliorate, we can apply the simplifying constrain of only allowing Bob and Alice to deviate for a single action.
% Let $\mathcal{D}(b)$ be the set of actions of player~1 that signal deviation, i.e., actions that cannot be played in this PBS under the blueprint, and let $\mathcal{R}$ be the set of all plausible responses, i.e. a set of actions that player~2 can play in any state.
% More formally,
% $$
% \mathcal{D}(b) := \{a_1 \in \mathcal{A}_1 | \forall s_1 \in b :\pi(a_1 | b, s_1) = 0 \}, \;\; \mathcal{R} := \bigcap_{s_2 \in \mathcal{S}_2}\mathcal{A}_2(s_2).
% $$
% Now we define expected value of a deviations $(a_1, a_2)$ as
% \begin{equation}
%      q_{\pi}(b, s_1, a_1, a_2) =
%     \begin{cases}
%       E_{h~b, h\in_s_1}, & \text{if}\ a_1 \in \mathcal{D}(b), a_2 \in \mathcal{R} \\
%       -\infty, & \text{otherwise}
%     \end{cases}
% \end{equation}
To ameliorate, we can apply the simplifying constraint of only allowing Alice and Bob to deviate for a single action pair regardless of Alice's information state $s_1$.
For simplicity, we also set $\epsilon = 0$. Let $\mathcal{D}(b)$ be the set of plausible deviation actions for Alice, i.e., action that cannot be played under blueprint: $\mathcal{D}(b) := \{a_1 \in \mathcal{A}_1 | \forall s_1 \in b :\pi(a_1 | b, s_1) = 0 \}.$ 
Let $\mathcal{R}$ be the set of all plausible actions that Bob can play in response in every possible state, i.e., $\mathcal{R} := \bigcap_{s_2 \in \mathcal{S}_2}\mathcal{A}_2(s_2)$. If $\mathcal{R} = \emptyset$, Alice will skip searching for SED this turn and follow BP.
Now we can augment the expected value of a deviation with the allowed range of actions:
\begin{equation}
     \hat{q}_{\pi}(b, s_1, a_1, a_2) =
    \begin{cases}
      q_{\pi}(b, s_1, a_1, a_2), & \text{if}\ a_1 \in \mathcal{D}(b), a_2 \in \mathcal{R}, \\
      -\infty, & \text{otherwise.}
    \end{cases}
\end{equation}
Then we can rewrite equation~(\ref{eq:general}) using our newly defined $q$ function as follows:
\begin{equation}
(a_1^{\ast}, a_2^{\ast}) = \argmax_{a_1, a_2} \mathbb{E}_{s_1 \sim b} \max[\hat{q}_{\pi}(b, s_1, a_1, a_2),  q_{\pi}(b, s_1)].
\end{equation}
These values could be found in $O(|\mathcal{S}_1| |\mathcal{A}_1| | \mathcal{A}_2|)$ time, given the two inner $q$ functions.
%, and in $O(|\mathcal{A}_1| | \mathcal{A}_2|)$ time, given $q_{\text{dev}}(a_1, a_2) = \mathbb{E}_{s_1 \sim b} \max(\hat{q}_{\pi}(b, s_1, a_1, a_2), q_{\pi}(b, s_1))$.
Moreover, as all of these depend only on the public belief state, both players can compute them independently, assuming that there are no ties.

Once the deviation pair is computed, Alice decides whether to proceed with the deviation or not given her private information: Alice plays $a_1^{\ast}$ if $\hat{q}_{\pi}(b, s_1, a_1^{\ast}, a_2^{\ast}) > q_{\pi}(b, s_1)$, and the blueprint action otherwise.
It is possible that $(a_1^{\ast}, a_2^{\ast})$ is not a valid deviation pair if no plausible deviations exist, e.g., $a_1^{\ast}$ could be outside of $\mathcal{D}(b)$.
However, in this case $\hat{q}_{\pi}(b, s_1, a_1^{\ast}, a_2^{\ast})$ is $-\infty$, and so Alice will resort to playing blueprint.
Bob can detect the deviation as $\mathcal{D}(b)$ is a public knowledge and $a_1^{\ast} \in \mathcal{D}(b)$.
Therefore, depending on his observation, he can either play $a_2^{\ast}$ or respond as usual.

% By replacing $\argmax$ with $\max$ in~\ref{eq:finesse} one can show that finesse has a weak policy improvement guarantee.

% Assuming that $\forall (a_1, a_2) \notin  (\mathcal{D}(b) \times \mathcal{R}): \:  q_{\pi}(b, s_1, a_1, a_2) = -\infty$, we can represent the value of the policy that allows a single deviation at $b$ and plays blueprint after that in the following simple way:

We can represent the value of the policy that allows a single deviation at $b$ and plays blueprint afterwards as: $q_\pi^{\ast}(b) = \max_{a_1, a_2} \mathbb{E}_{s_1 \sim b} \max(\hat{q}_{\pi}[b, s_1, a_1, a_2), q_{\pi}(b, s_1)]$.
It is easy to see that $q_\pi^{\ast} \geq q_\pi$, i.e., this algorithm has a weak policy improvement guarantee.

If we apply this procedure to the tiger-trampoline game with a blueprint policy that always no-ops:
% {\small
\begin{align*}
q_\pi^{\ast}(b) &=\max_{a_1, a_2} \mathbb{E}_{s_1 \sim b} \max[\hat{q}_{\pi}(b, s_1, a_1, a_2), q_{\pi}(b, s_1)] \\
& = \max_{a_1, a_2} [P(\mbox{tiger}) \max(\hat{q}_{\pi}(b, \mbox{tiger}, a_1, a_2), 0) + P(\mbox{trampoline}) \max(\hat{q}_{\pi}(b, \mbox{trampoline}, a_1, a_2), 0)]\\
&= P(\mbox{trampoline}).
\end{align*}
As expected, we obtain a policy improvement with value equal to the probability of a trampoline. We discover the SED where Alice jumps when a trampoline is present and Bob pulls the lever. We provide a proof-of-principle implementation of \ourmethod~in the trampoline-tiger game that can be run online at \url{https://bit.ly/3KtMLT6}.

% \section{Learning Finesses}

% Is it possible to learn to finesse in a model-free setting?
% Given a mechanism to compactify the belief state, the answer is yes.
% First consider that the functions $q_{\pi}(b, s_1, a_1, a_2)$ and $q_{\pi}(b, s_1)$ can be learned from samples, for any fixed $s_1, a_1, a_2$.
% Now, given that these functions are learned, it is simple to compute $\max(q_{\pi}(b, s_1, a_1, a_2), q_{\pi}(b, s_1))$.
% Lastly, we can compute 
% \[\mathbb{E}_{s_1 \sim b} \max(q_{\pi}(b, s_1, a_1, a_2), q_{\pi}(b, s_1))\]
% for a fixed $a_1, a_2$.
% Doing this for each $a_1, a_2$ yields the function over which we need to argmax.

% One possibile use of learning finesses is to bootstrap a policy by iteratively (i) adding finesses and (ii) folding that finesse into the blueprint policy.
% This procedure would probably require beginning with a reasonable blueprint policy.

\subsection{Coordination by Extending Conventions}

In a coordination context, we cannot require Bob to perform arbitrary symmetry breaking, i.e., choose an action among several with the same expected value.
For example, consider a case in which a set of multiple action pairs tied for having the maximal value
$\{(a_1=x, a_2=y), (a_1=x, a_2=z), (a_1=w, a_2=z)\}$.
% In this case, Alice can decide to break symmetry arbitrarily among $x$ and $w$---in either case, Bob will know that Alice is finessing.
Say that Alice picks $a_1 = x$.
This is problematic for Bob because he has no information about whether Alice is selecting $x$ because she is in an information state in which $(x, y)$ is good or whether she is in an information state in which $(x, z)$ is good (the states for which they are good may be disjoint).

One way to resolve this issue is to force Alice to respect Bob's inabilities to break ties by making the response function stochastic, e.g., a softmax over the expected values.
That is, given that Alice deviates with $a_1$, Alice assumes that Bob plays
\begin{equation}
a_2^{\ast} \sim \softmax_{a_2} \left[\mathbb{E}_{s_1 \sim b} \max[\hat{q}_{\pi}(b, s_1, a_1, a_2), q_{\pi}(b, s_1)] / t \right]
\end{equation}
with a temperature hyper-parameter to control the sharpness of the distribution. Then Alice chooses to deviate using 
\begin{equation}
\argmax_{a_1} \mathbb{E}_{s_1 \sim b} \max[\mathbb{E}_{a_2 \sim a_2^{\ast}(b, a_1)} \hat{q}_{\pi}(b, s_1, a_1, a_2), q_{\pi}(b, s_1)].
\end{equation}
Bob can either play according to the softmax or select the single maximum, as Alice has already picked the deviation assuming that Bob cannot break symmetries. 
This is optimal if Alice can only play one single deviation action across \emph{all her information states}. 
In the following we'll address how to deal with multiple deviations, which also makes the formulation easier to understand. 

\subsection{Taking Alice's Information State Into Account}
Our discussion above has largely ignored the fact that Alice can perform different deviations given her information state $s_1$ (i.e., private observation). 
Nonetheless, it helps us find the response function $f(b, a_1) = \softmax_{a_2} \left[\mathbb{E}_{s_1 \sim b} \max[\hat{q}_{\pi}(b, s_1, a_1, a_2), q_{\pi}(b, s_1)] / t\right]$. 
Given that Alice knows Bob will response according to $a_2 \sim f(b, a_1)$, Alice is now free to optimize her action by taking her information state back into consideration.
By applying Jensen's inequality, we observe that
\begin{align*}
&\max_{a_1} \mathbb{E}_{s_1 \sim b} \max\left[\mathbb{E}_{a_2 \sim f(b, a_1)} \hat{q}_{\pi}(b, s_1, a_1, a_2), q_{\pi}(b, s_1)\right] \\
&\leq \mathbb{E}_{s_1 \sim b}  \max_{a_1} \max\left[ \mathbb{E}_{a_2 \sim f(b, a_1)} \hat{q}_{\pi}(b, s_1, a_1, a_2), q_{\pi}(b, s_1)\right].
\end{align*}
Therefore, at information state $s_1$, Alice opts out of the blueprint if the \emph{optimal deviation action}, $a_1^* =\max_{a_1} \mathbb{E}_{a_2 \sim f(b, a_1)} \hat{q}_{\pi}(b, s_1, a_1, a_2)$, exceeds the expected return of the BP, $ q_{\pi}(b, s_1)$. This is the final formulation of IMPROVISED$^E$. Please refer to the Appendix~\ref{sec:pseudocode} for the detailed pseudocode.

We also define IMPROVISED$^P$, where Bob plays according to the probability of improvement 
\begin{align}
a_2^{\ast} \sim \softmax_{a_2} \left[\mathbb{E}_{s_1 \sim b} \, I[\hat{q}_{\pi}(b, s_1, a_1, a_2) > q_{\pi}(b, s_1)] / t\right]
\end{align}
where $I$ is the indicator function and Alice selects any action maximizing the expected improvement.
While IMPROVISED$^P$ does not maximize the expected return as IMPROVISED$^E$ does, it may better reflect human SEDs since it finds deviation pairs that maximize the \emph{probability of improvement}.

%% file: 5_experiment.tex
\section{Experiments}

\begin{wrapfigure}{r}{0.45\textwidth}
    \vspace{-9mm}
    \centering
    \includegraphics[width=.45\textwidth]{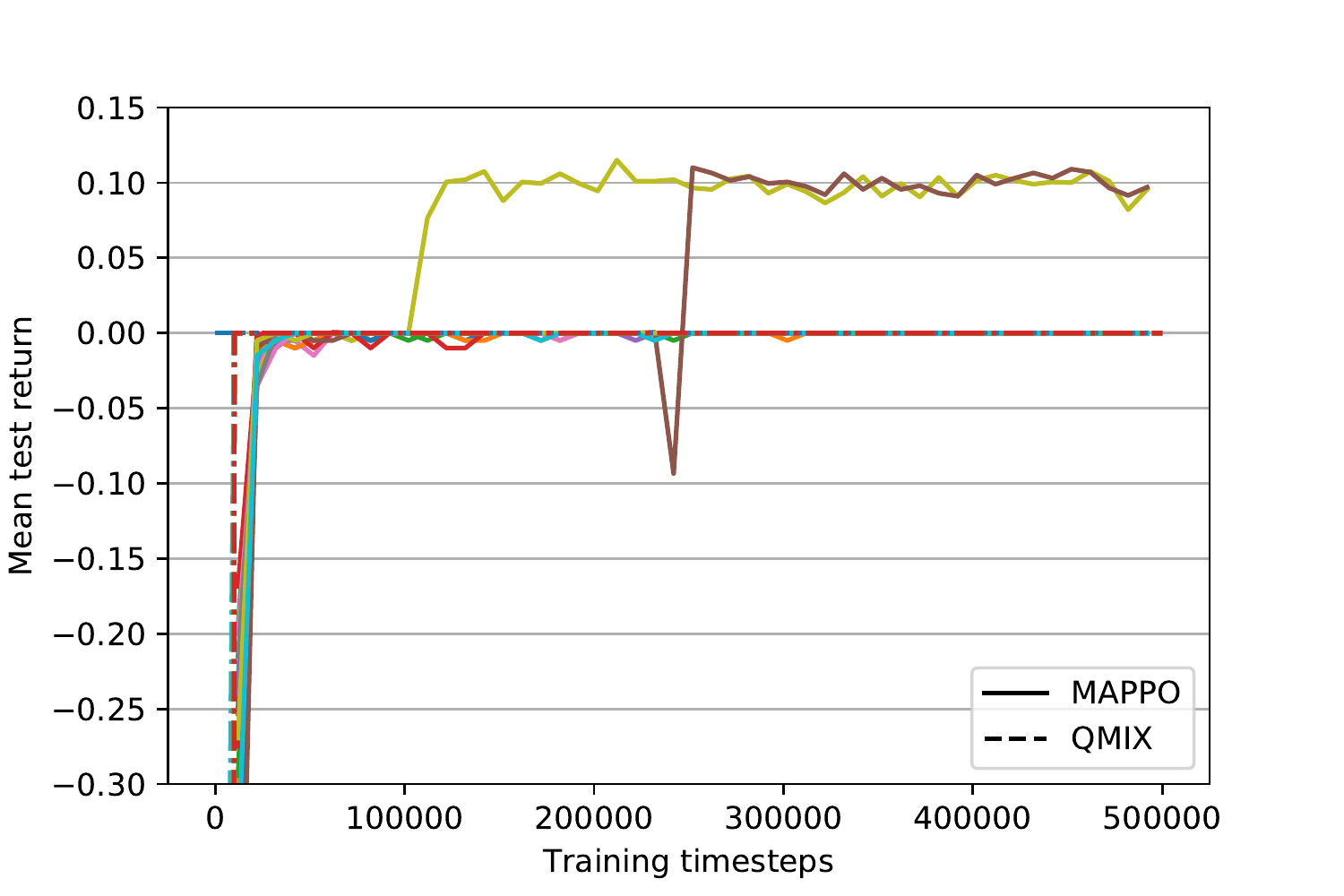}
    \caption{\small 20 independent runs of MAPPO (solid lines) and 24 of QMIX (dashed lines) on the tiger-trampoline toy problem. 2 MAPPO runs find the optimal strategy while no QMIX run does.}
    \vspace{-9mm}
    \label{fig:toy-result}
\end{wrapfigure}

We test the IMPROVISED in two different settings. The first setting is the trampoline-tiger game explained before. Secondly, we apply IMPROVISED to three-player Hanabi, where we start from a BP trained on human data. 

\subsection{Trampoline Tiger}
\label{sec:trampolineresults}

As illustrated in Section~\ref{sec:easier}, in the trampoline-tiger game  \ourmethod$^E$ and \ourmethod$^P$ both recover the optimal SED, when starting from the no-op BP in $100\%$ of the runs. Alice will decide to jump whenever there is a trampoline and Bob then pulls the lever to open the door, leading to the optimal expected return of 0.1.

For comparison, we also ran 20 unique seeds of MAPPO
% \footnote{We used the default parameters from \url{https://github.com/marlbenchmark/on-policy} used for Hanabi, except with a hidden size of 128, and reducing the number of threads by a factor of 2.} \citep{yu2021surprising}
and 24 different hyper-parameter combinations of QMIX \citep{rashid2018qmix} on the trampoline-tiger problem. The results are shown in Figure~\ref{fig:toy-result}. All runs converged extremely rapidly to a policy that avoids the highly negative payoffs, but only 2/20 of the MAPPO runs and 0/24 of the QMIX runs discovered the optimal strategy within 500k timesteps, leading to an average reward across runs of 0.01 for MAPPO and 0.0 for QMIX. The code and hyper-parameters are available in the supplementary material.

In our experiments, a variety of other standard multi-agent learning techniques could not solve this toy problem at all, including value-decomposition network~\cite{sunehag_value-decomposition_2018} and simplified action decoder~\cite{sad}.

\subsection{Hanabi}
\label{sec:hanabi-def}
Hanabi has been proposed as a benchmark challenge for multi-agent coordination, and has been the subject of significant research \citep{hanabi, hu2020other}. Briefly, in Hanabi, there is a a 50 card deck consisting of 5 different suits (colors) and 5 rank of card within each suit, for a total of 25 unique combinations, with some duplicates. 2 to 5 players cooperatively take turns playing cards, giving hints to other players, or discarding. Players may see their partners' hands, but \emph{not} their own. The team's goal is to play exactly one card of each rank in each suit in increasing order, scoring 1 point per successful play. Upon three plays of cards duplicated or out-of-order, the players lose and instead score 0. Giving a hint spends one of 8 shared hint tokens, allowing one to name a color or rank in a partner's hand and indicate all cards of that color or rank in that hand. Discarding a card replenishes one hint token. See also \cite{hanabi} for a more thorough description of the rules and basic strategy.

\subsection{What is a Finesse?}
\label{sec:whatisfinesse}
As mentioned earlier in Section \ref{sec:sedsbackground}, a clear example of a SED is the finesse move in Hanabi. Although described in far more detail with an example in \citep{hanabi}, we will recap it briefly here. For convenience, we always refer to the current active player of a given turn as {\it player 1} and the players who move subsequently as {\it player 2}, {\it player 3} etc.

The standard, vanilla form of the finesse in Hanabi occurs when {\it player 1} gives a hint to {\it player 3} (i.e. the player who moves two steps later) that under the conventions common between the players implies that player {\it player 3} should play a particular card. Unbeknownst to {\it player 3}, that card, if played, would be out-of-order and would fail disastrously. Instead, {\it player 2} has just drawn the card that is both playable and if played makes {\it player 3}'s card playable next. {\it Player 2} is expected to realize that the only way that {\it player 1}'s hint could be good rather than a disaster is if  {\it player 2} themself has newly drawn this exact card, and to take a leap of faith and play that newly drawn card blindly, having no explicit information about it. The end result is that {\it player 1} signals in only one action for both other players to each successfully play a card, a large gain.

While other finesse-like patterns are possible, in our Hanabi experiments for simplicity we focus only those finesse moves that follow the pattern described above. 

\subsection{IMPROVISED in Hanabi}
% \hengyuan{write overview of the experiments. 1) use finesse to check the whether our algorithm performs SEDs. 2) check if SEDs bring extra benefits in terms of the expected return.}

\begin{table*}
\begin{subtable}[c]{0.6\textwidth}
\centering
    \begin{tabular}{l c r r}
        \toprule
        Model & Finesse-able & \multicolumn{2}{}{}{Finesse-complete} \\
        \midrule
        SAD~\cite{sad}              &   1375 & 597  (43.42\%) \\
        Other-Play~\cite{otherplay} &   1537 & 512  (33.31\%) \\
        OBL (Level 5)~\cite{obl}    &   1100 & 356  (32.36\%) \\
        Behavior Clone~\cite{obl}        &   1376 & 1183 (85.97\%) \\
    \bottomrule    
    \end{tabular}
    \subcaption{\small Number of finesse-able situations over 1000 games and \% of those where {\it player 3} will correctly respond to complete the finesse if {\it player 1} and {\it 2} were forced to play the finesse move. The Behavior Clone model has a much higher chance to complete a finesse. }
    \label{tab:hanabi-bp}
\end{subtable}
\hfill
\begin{subtable}[c]{0.35\textwidth}
\centering
  \begin{tabular}{l c}
        \toprule
        Method  & Finesse \\
        \midrule
        Blueprint &  0       \\
        SPARTA~\cite{sparta} &  15         \\
        IMPROVISED$^E$ &  51     \\
        IMPROVISED$^P$ &  62     \\
    \bottomrule
    \end{tabular} 
    \subcaption{\small Number of finesses executed by different methods at 200 finesse-complete situations using Behavior Clone agent from Table~\ref{tab:hanabi-bp} as blueprint.}
    \label{tab:hanabi-finesse-one-step}
\end{subtable}
\caption{IMPROVISED for Finesse in Hanabi}
\end{table*}

We design experiments to evaluate IMPROVISED in Hanabi from both qualitative and quantitative perspectives. In the qualitative evaluation, we study whether IMPROVISED can indeed perform finesses in manually selected situations where finesse could be completed (finesse-complete situations). 
We are particularly interested in the finesse style SEDs as they are most natural and intuitive way to demonstrate IMPROVISED's ability to perform SEDs.
Note that IMPROVISED discovers many types of SEDs in Hanabi beyond the finesse style moves because it searches for any beneficial joint deviations that when {\it player 1} initiates a deviation, {\it player 2} can independently figure out the correct deviation in response. 
However, other types of SEDs may be hard for human to interpret and less intuitive to analyse and present.
In the quantitative evaluation, we show that IMPROVISED as a planning algorithm improves the expected return when applied to test finesse-able situations and entire games.

To implement IMPROVISED in Hanabi, we first need a belief function from which we can sample game states given either public or private knowledge of the game to perform Monte Carlo rollouts. Luckily, the belief over possible hands in Hanabi can be computed analytically~\cite{sparta}. The belief is first initialized to cover all possible hands and then incrementally updated by filtering out hands that contradict with public knowledge revealed through hints and that would have caused the players to pick different moves at each time step. Since we are mainly concerned with the finesse style SEDs in the experiments, we restrict the search action space $\mathcal{A}_1$ to contain only the hint moves that target {\it player 3} and $\mathcal{A}_2$ to contain only the play moves. {\it Player 1} and {\it player 2} compute their own copies of the response function $f$ independently. The detailed hyper-parameters and computational cost are in the Appendix.
% should we add more details?

To check IMPROVISED's ability to perform finesse, we first generate situations where seasoned human players may carry out finesse moves. 
We use a blueprint policy $\pi$ to generate selfplay games over a range of decks (game seeds) and look for situations where {\it player 1} observes that {\it player 2}'s newest draw card is playable and has not been hinted at, and meanwhile {\it player 3} holds an un-hinted card that can be played after {\it player 2} plays their newest card (finesse-able situations). 
Then we manually override {\it player 1}'s and {\it player 2}'s moves in finesse-able situations to carry out the finesse and check if {\it player 3} will play the designated card under the blueprint to make the finesse complete and worthwhile (finesse-complete situations). 
We experiment with 4 different blueprints and show their statistics of finesse-complete situations over 1000 game seeds in Table~\ref{tab:hanabi-bp}. 
All the agents are trained following the settings in corresponding prior works. Although finesse-able situations are pretty common in the games played by all four agents, the first three RL agents trained without human data rarely complete a manually enforced finesse move, making them undesirable for the subsequent experiment. It also worth noting that none of the RL blueprints perform finesses by themselves while the behavior clone agent performs only 1 finesse out of the 1183 finesse-complete situations.

We apply both IMPROVISED$^E$ and IMPROVISED$^P$ with the behavior clone blueprint on 200 randomly chosen finesse-complete situations to check how many finesse it performs. The results are shown in Table~\ref{tab:hanabi-finesse-one-step}. 
Due to the novelty of this problem setting, there has been no prior method to directly compare against. For reference, we run SPARTA~\cite{sparta}, a strong search algorithm for Dec-POMDP designed to find unilateral deviations that maximizes the expect return, on the same situations with the same restrictions on the set of actions that the agents may deviate to.
From the table, we see that both IMPROVISED algorithms perform significantly more finesses, indicating their effectiveness in finding SEDs on the fly. In the situations where IMPROVISED does not perform finesse, it can either be that IMPROVISED finds no beneficial deviations or it finds better, non-finesse SEDs. It is also interesting to see that IMPROVISED$^P$ finds more finesses than IMPROVISED$^E$, which is in line with our hypothesis that humans might use the \emph{probability of improvement} when reasoning about SEDs instead of relying on small difference between different (joint) q-values. 

Then, we run IMPROVISED on the full game of Hanabi. At any time step, the active player will first check whether a deviation has been initiated by other players by checking whether previous players have picked actions with low probability under the blueprint. 
If a deviation has been initiated, they will play their corresponding role as either {\it player 2} or {\it player 3}. 
Otherwise the active player will decide whether to deviate using the IMPROVISED method. 
% To reduce the computation cost we reduce $M$ to $400$ and share the result of $a_2^*(a_1)$ between Player 1 and Player 2 instead of computing it twice independently as we empirically find that the statistic is stable enough against random seeds. A full game then takes around 10-12 hours using 20 CPU cores and 2 GPUs. 
Over a fixed set of 100 game seeds, IMPROVISED$^E$ performs \textbf{29} finesses while SPARTA only performs 3. Although it is still much less frequent than what expert human players will do, IMPROVISED is nonetheless a meaningful step along this novel and challenging direction.

Although our work is focused on understanding and replicating this human capability rather than optimizing for self-play score, we also report quantitative results showing a noticeable increase in score using IMPROVISED compare to the plain blueprint, giving evidence that the deviations found do have a meaningful impact on the game. 
If we run IMPROVISED only at the 200 finesse-complete situations from Table~\ref{tab:hanabi-finesse-one-step} and use blueprint for the rest of the timesteps, then IMPROVISED$^E$ and IMPROVISED$^P$ achieve average scores of 18.58 and 18.36 respectively while then blueprint gets 17.52 on those same games. 
When we run IMPROVISED$^E$ on the full game, we improve the average score from 17.8 to 23.54.

%% file: 6_related_work.tex
\section{Related Work}

\subsection{Public Belief Methods} 
Our method builds on prior work that models Dec-POMDs as PuB-MDPs~\cite{nayyar,oliehoek2013sufficient,dibangoye2016optimally,foerster_bayesian_2019,sokota2021solving,bft}. More recently, this same approach has been extended to two-player zero-sum games~\cite{moravvcik2017deepstack,kovavrik2019rethinking,brown2020combining,kartik2020upper,buffet2020bellman} and two-team zero-sum games~\cite{kartik2021common,2team0s}. PuB-MDPs enable theoretically sound planning algorithms without having to reason about the agents' entire policies because the public belief state serves as a sufficient statistic for planning. However, prior work on PuB-MDPs do not explicitly search for SEDs.

\subsection{Search in Dec-POMDPs} 
Search and planning algorithms have been a critical component to AI breakthroughs in complex multi-agent environments, including backgammon~\cite{tesauro1995temporal}, chess~\cite{campbell2002deep}, go~\cite{silver2016mastering,silver2017mastering,silver2018general}, and poker~\cite{moravvcik2017deepstack,brown2017superhuman,brown2019superhuman}. Closely related to our work are methods for policy improvement via planning in Dec-POMDPs, such as SPARTA~\citep{sparta}. SPARTA conducts a one-step lookahead and chooses the action that maximizes expected value assuming all players play according to a common-knowledge BP thereafter. However, SPARTA searches only for unilateral deviations from the BP rather than multilateral deviations, because searching over all of the latter would be intractable. As discussed in Section \ref{sec:examiningseds}, discovering SEDs may require considering multi-lateral deviations from the BP. IMPROVISED is able to discover these multi-lateral deviations by searching over a constrained set of multi-lateral deviations. Other prior work has also investigated planning algorithms for Dec-POMDPs~\citep{katt2017learning}. However, most of these techniques have typically searched over public belief states~\citep{dibangoye2016optimally,fehr2018rho,sokota2021solving}. This greatly increases the complexity of the search procedure.

%% file: 7_conclusion.tex
\section{Conclusions}

The human ability to coordinate with others using minimal explicit agreement and extending conventions ``on the fly'' is one of the most intriguing reasoning abilities and far ahead of any of our learning methods. In this paper we formalize the definition of such behavior as \emph{self-explaining deviations}.
We show why existing methods fail to learn these types of deviations and presented an optimal algorithm named IMPROVISED that can both discover them and correspond correctly when they are being carried out by other agents at test time. Crucially, \ourmethod~allows agents to improve performance at test time by improving upon a common knowledge blueprint policy, even when the blueprint is a Nash equilibrium. 

\textbf{Limitation and Future Work} In this paper we first address SED with planning without taking into account the extra complexity of learning and approximating due to its difficulty and novelty. The resulted algorithm is computationally expensive as it requires Monte Carlo rollouts to estimate many crucial quantities at test time. 
However, it is possible to derive a more efficient, learning based version of IMPROVISED since the joint Q-function $q(b, s_1, a_1, a_2)$ can be learned. 
There are challenges to make these estimates accurate on the deviating states but our paper nonetheless provides a valuable stepping stone along this exciting future direction.

\textbf{Societal Impact} One of the most fundamental goal of multi-agent learning is to produce agents that collaborate well with human in real world. Therefore, it is important for the artificial agents to not only understand human conventions, but also decode human's intention when deviations occur so that agents can response correctly. Such agents should also be capable to initiate such self-explaining deviations when beneficial to maximize their social values. Therefore, this direction of work to understand and replicate human's ability to perform SEDs using theory of mind is important and positive to the society. We currently do not foresee any negative societal impact from this work.

%% file: A_algo.tex
\section{Pseudocode of IMPROVISED$^{E}$}
\label{sec:pseudocode}

\begin{algorithm}[H]
\small
% \SetAlgoLined
\caption{IMPROVISED$^E$}
% \KwResult{}
\begin{algorithmic}
\STATE \textbf{Definitions:} \\
% \STATE \textbullet~$a_1^{bp}$
\STATE \textbullet~$b$: common public belief of player $P_1$ and player $P_2$ 
\STATE \textbullet~$\mathcal{A}_i$: action space of $P_i$ \\
\STATE \textbullet~$s_i$: information state of $P_i$ \\
% \textbullet~$A_i(s_i)$: legal action of $P_i$ in $s_i$ \\
\STATE \textbullet~$b(s_1)$: belief of $P_2$ given $P_1$'s information state $s_1$ \\
\STATE \textbullet~$\pi$: joint blueprint policy \\
\STATE \textbullet~$R(s_1$, $s_2$, $\pi$, $[a_1, a_2])$: reset current game state with $s_1$, $s_2$, rollout until termination following (the optional $[a_1, a_2]$ and then) $\pi$, and return the total reward. \\
\STATE \textbf{Method:} \\
\STATE initialize $q_{\pi}(a_1, a_2, b) = 0$ for  $(a_1, a_2) \in \mathcal{A}_1 \times \mathcal{A}_2$
\STATE sample $M$ private state for $P_1$, $s^{(1)}_1, \dots, s_1^{(M)} \sim b$ \\
\STATE $P_{\pi}(a_1) = \frac{1}{M}\sum_{i=1}^{M}\pi(a_1 |b(s_1^{(i)}))$ for $a_1 \in A_1$ \\
\FOR{$s_1^{(i)} \in s^{(1)}_1, \dots, s_1^{(M)}$}
    \STATE sample $N$ private state for $P_2$, $s_2^{(1)}, \dots, s_2^{(N)} \sim b(s_1^{(i)})$ \\
    \STATE $q_{\pi}(b, s_1^{(i)}) = \frac{1}{N}\sum_{j} R(s_1^{(i)}, s_2^{(j)}, \pi)$ \\
    \FOR{$(a_1, a_2) \in \mathcal{A}_1 \times \mathcal{A}_2 $}
        \IF{$ P_{\pi}(a_1) \geq \epsilon_p $}
        \STATE $q_{\pi}(a_1, a_2, b, s_1^{(i)}) = -\infty$
        \ELSE
        \STATE $q_{\pi}(a_1, a_2, b, s_1^{(i)})= \frac{1}{N}\sum_j R(s_1^{(i)}, s_2^{(j)}, \pi, a_1, a_2)$ \\
        \ENDIF 
        % \STATE $q(a_1, a_2, b) \mathrel{+}= \max[q(a_1, a_2, b, s_1^{(i)}), q_{\pi}(b, s_1^{(i)})]$\\
    \ENDFOR
\ENDFOR
\FOR{$(a_1, a_2) \in \mathcal{A}_1 \times \mathcal{A}_2$}
\STATE $q_{\pi}(a_1, a_2, b) = \frac{1}{M} \sum_{i}\max\left[q_{\pi}(a_1, a_2, b, s_1^{(i)}), q_{\pi}(b, s^{(i)}_1) \right]$ %for $(a_1, a_2) \in A_1 \times A_2$ \\
\ENDFOR
\FOR{$a_1 \in A_1$}
\STATE $f(b, a_1) = \softmax_{a_2} \left[ q_{\pi}(a_1, a_2, b) / t \right]$  \\
\STATE $q_{\pi}(b, s_1, a_1) = \mathbb{E}_{s_2' \sim b(s_1), a_2 \sim f(b, a_1)} R(s_1, s_2', \pi, a_1, a_2)$
\ENDFOR
\IF{$\max q_{\pi}(b, s_1, a_1) \geq q_{\pi}(b, s_1) + \epsilon_q$}
\STATE \textbf{return} $\argmax_{a_1}q_{\pi}(b, s_1, a_1)$ \\
\ELSE
\STATE \textbf{return} $a_1^{bp}$ // the action under blueprint\\
\ENDIF
\end{algorithmic}
\label{algorithm:improvised_e}
\end{algorithm}

\section{Experimental Details for Tiger-Trampoline}

\begin{table}[h]
    \centering
\begin{tabular}{ c|c }
\toprule
Hyper-parameter & Values \\
\midrule
learning rate            & 0.0005, 0.0001 \\
batch size               & 16, 32         \\
$\varepsilon$ annealing period & 20000, 10000   \\
RNN hidden dimension     & 64, 32, 16    \\
\bottomrule
\end{tabular}
\caption{Hyper-parameters of QMIX in the Tiger-Trampoline Experiment}
\label{tab:qmix-hparam}
\end{table}

In Section~\ref{sec:trampolineresults}, we show the results of MAPPO and QMIX on the Tiger-Trampoline game. For the MAPPO we use the default parameters from the open sourced implementation\footnote{ https://github. com/marlbenchmark/on-policy} used for Hanabi, except with a hidden size of 128, reducing the episode length cap, and reducing the number of threads by a factor of 2. For QMIX, we use the open sourced implementation\footnote{https://github.com/oxwhirl/pymarl} of the algorithm provided as part of the PyMARL framework~\cite{samvelyan19smac}. We used the default agent and training configuration, except for the four hyper-parameters listed in table~\ref{tab:qmix-hparam}. For those, we tried all combinations of the corresponding values, producing a total of 24 runs, each training for 500k steps, or 250k episodes.

\section{Experimental Details for Finesse in Hanabi}

In the Hanabi experiments, we implement IMPROVISED as follows (better viewed together with the pseudocode). The belief $b$ is the common public belief shared by {\it player 1} and {\it player 2} based on common knowledge available to all players and their common private knowledge of {\it player 3}'s hand. 
We first draw $M$ Player 2 hands $s_1'$ from $b$ and compute blueprint actions $a_{\pi} = \pi(b(s_1'))$ and $P_{\pi}(a)$. 
We then consider joint actions $\mathcal{A}_{1} \times \mathcal{A}_{2} = \{(a_1, a_2) | P_{\pi}(a_1) \leq 0\}$ for {\it player 1} and {\it player 2}. 
Since our goal is to find finesse style joint deviations, we further restrict $a_1$ to be a \textit{hint move} to {\it player 3} and $a_2$ to be a \textit{play move}. 
Given $s_1'$, {\it player 1} can further induce the private belief $b(s_1')$ over their own hand. 
For each of $s_1'$, {\it player 1} calculates Monte Carlo estimations of $q(a_1, a_2, b, s_1',)$ for $(a_1, a_2) \in \mathcal{A}_{1} \times \mathcal{A}_{2}$ and $q_{\pi}(b, s_1')$ with $N$ samples drawn from $b(s_1')$. 
So far we have collected all the quantities required to compute the mapping $f$ for IMPROVISED$^{P}$ and for IMPROVISED$^{E}$. 
Finally, we draw another $K$ samples from the true $b(s_1)$ where $s_1$ now is the real hand of {\it player 2} to estimate $\delta = \max_{a_1} \mathbb{E}_{a_2 \sim f(b, a_1)} q_{\pi}(b, s_1, a_1) - q_{\pi}(b, s_1)$. {\it Player 1} will deviate to $\argmax_{a_1} \mathbb{E}_{a_2 \sim a_2^{\ast}(a_1)} q_{\pi}(b, s_1, a_1, a_2)$ if $\delta \geq 0.05$. In the next turn, {\it player 2} can carry out the same computation process to get $P_\pi(a_1)$ and $f(b, a_1)$ to figure out whether {\it player 1} has deviated and if so what is the correct response. {\it Player 1} and {\it player 2} do not share the random seed beforehand.

In the experiments where we run IMPROVISED on finesse-complete situations only, we set $M=1000$, $N=100$ and $K = 10000 / |\mathcal{A}_1|$. 
It takes roughly 2 hours in total for both {\it player 1} and {\it player 2} to compute the deviations independently using 5 CPU cores and 1 GPU. 

In the experiments where we run IMPROVISED on the full game of Hanabi, we reduce $M$ to $400$ and share the result of $f(b, a_1)$ between Player 1 and Player 2 instead of computing it twice independently as we empirically find that the statistic is stable enough against random seeds. A full game then takes around 10-12 hours using 20 CPU cores and 2 GPUs.

%% file: main.bbl
\begin{thebibliography}{35}
\providecommand{\natexlab}[1]{#1}
\providecommand{\url}[1]{\texttt{#1}}
\expandafter\ifx\csname urlstyle\endcsname\relax
  \providecommand{\doi}[1]{doi: #1}\else
  \providecommand{\doi}{doi: \begingroup \urlstyle{rm}\Url}\fi

\bibitem[Bard et~al.(2020)Bard, Foerster, Chandar, Burch, Lanctot, Song,
  Parisotto, Dumoulin, Moitra, Hughes, Dunning, Mourad, Larochelle, Bellemare,
  and Bowling]{hanabi}
N.~Bard, J.~N. Foerster, S.~Chandar, N.~Burch, M.~Lanctot, H.~F. Song,
  E.~Parisotto, V.~Dumoulin, S.~Moitra, E.~Hughes, I.~Dunning, S.~Mourad,
  H.~Larochelle, M.~G. Bellemare, and M.~Bowling.
\newblock The hanabi challenge: A new frontier for ai research.
\newblock \emph{Artificial Intelligence}, 280, 2020.
\newblock URL \url{https://doi.org/10.1016/j.artint.2019.103216}.

\bibitem[Brown and Sandholm(2017)]{brown2017superhuman}
N.~Brown and T.~Sandholm.
\newblock Superhuman {A}{I} for heads-up no-limit poker: Libratus beats top
  professionals.
\newblock \emph{Science}, page eaao1733, 2017.

\bibitem[Brown and Sandholm(2019)]{brown2019superhuman}
N.~Brown and T.~Sandholm.
\newblock Superhuman {A}{I} for multiplayer poker.
\newblock \emph{Science}, page eaay2400, 2019.

\bibitem[Brown et~al.(2020)Brown, Bakhtin, Lerer, and Gong]{brown2020combining}
N.~Brown, A.~Bakhtin, A.~Lerer, and Q.~Gong.
\newblock Combining deep reinforcement learning and search for
  imperfect-information games.
\newblock In H.~Larochelle, M.~Ranzato, R.~Hadsell, M.~F. Balcan, and H.~Lin,
  editors, \emph{Advances in Neural Information Processing Systems}, volume~33,
  pages 17057--17069. Curran Associates, Inc., 2020.
\newblock URL
  \url{https://proceedings.neurips.cc/paper/2020/file/c61f571dbd2fb949d3fe5ae1608dd48b-Paper.pdf}.

\bibitem[Buffet et~al.(2020)Buffet, Dibangoye, Delage, Saffidine, and
  Thomas]{buffet2020bellman}
O.~Buffet, J.~Dibangoye, A.~Delage, A.~Saffidine, and V.~Thomas.
\newblock On bellman's optimality principle for zs-posgs.
\newblock \emph{arXiv preprint arXiv:2006.16395}, 2020.

\bibitem[Campbell et~al.(2002)Campbell, Hoane~Jr, and Hsu]{campbell2002deep}
M.~Campbell, A.~J. Hoane~Jr, and F.-h. Hsu.
\newblock Deep blue.
\newblock \emph{Artificial intelligence}, 134\penalty0 (1-2):\penalty0 57--83,
  2002.

\bibitem[Dibangoye et~al.(2016)Dibangoye, Amato, Buffet, and
  Charpillet]{dibangoye2016optimally}
J.~S. Dibangoye, C.~Amato, O.~Buffet, and F.~Charpillet.
\newblock Optimally solving dec-pomdps as continuous-state mdps.
\newblock \emph{Journal of Artificial Intelligence Research}, 55:\penalty0
  443--497, 2016.

\bibitem[Fehr et~al.(2018)Fehr, Buffet, Thomas, and Dibangoye]{fehr2018rho}
M.~Fehr, O.~Buffet, V.~Thomas, and J.~Dibangoye.
\newblock $\rho$-pomdps have lipschitz-continuous $\varepsilon$-optimal value
  functions.
\newblock In \emph{Proceedings of the 32nd International Conference on Neural
  Information Processing Systems}, pages 6933--6943, 2018.

\bibitem[Foerster et~al.(2019)Foerster, Song, Hughes, Burch, Dunning, Whiteson,
  Botvinick, and Bowling]{foerster_bayesian_2019}
J.~N. Foerster, H.~F. Song, E.~Hughes, N.~Burch, I.~Dunning, S.~Whiteson,
  M.~Botvinick, and M.~Bowling.
\newblock {Bayesian Action Decoder for Deep Multi-Agent Reinforcement
  Learning}.
\newblock In \emph{{Proceedings of the 36th International Conference on Machine
  Learning}}, pages 1942--1951. {PMLR}, 2019.

\bibitem[Hu and Foerster(2020)]{sad}
H.~Hu and J.~N. Foerster.
\newblock Simplified action decoder for deep multi-agent reinforcement
  learning.
\newblock In \emph{International Conference on Learning Representations}, 2020.
\newblock URL \url{https://openreview.net/forum?id=B1xm3RVtwB}.

\bibitem[Hu et~al.(2020{\natexlab{a}})Hu, Lerer, Peysakhovich, and
  Foerster]{otherplay}
H.~Hu, A.~Lerer, A.~Peysakhovich, and J.~Foerster.
\newblock “{O}ther-play” for zero-shot coordination.
\newblock In H.~D. III and A.~Singh, editors, \emph{Proceedings of the 37th
  International Conference on Machine Learning}, volume 119 of
  \emph{Proceedings of Machine Learning Research}, pages 4399--4410. PMLR,
  13--18 Jul 2020{\natexlab{a}}.

\bibitem[Hu et~al.(2020{\natexlab{b}})Hu, Peysakhovich, Lerer, and
  Foerster]{hu2020other}
H.~Hu, A.~Peysakhovich, A.~Lerer, and J.~Foerster.
\newblock \textquotedblleft other-play\textquotedblright for zero-shot
  coordination.
\newblock In \emph{Proceedings of Machine Learning and Systems 2020}, pages
  9396--9407. 2020{\natexlab{b}}.

\bibitem[Hu et~al.(2021)Hu, Lerer, Cui, Pineda, Wu, Brown, and Foerster]{obl}
H.~Hu, A.~Lerer, B.~Cui, L.~Pineda, D.~Wu, N.~Brown, and J.~N. Foerster.
\newblock Off-belief learning.
\newblock \emph{(To Appear) ICML}, 2021.
\newblock URL \url{https://arxiv.org/abs/2103.04000}.

\bibitem[Kartik and Nayyar(2020)]{kartik2020upper}
D.~Kartik and A.~Nayyar.
\newblock Upper and lower values in zero-sum stochastic games with asymmetric
  information.
\newblock \emph{Dynamic Games and Applications}, pages 1--26, 2020.

\bibitem[Kartik et~al.(2021)Kartik, Nayyar, and Mitra]{kartik2021common}
D.~Kartik, A.~Nayyar, and U.~Mitra.
\newblock Common information belief based dynamic programs for stochastic
  zero-sum games with competing teams.
\newblock \emph{arXiv preprint arXiv:2102.05838}, 2021.

\bibitem[Katt et~al.(2017)Katt, Oliehoek, and Amato]{katt2017learning}
S.~Katt, F.~A. Oliehoek, and C.~Amato.
\newblock Learning in pomdps with monte carlo tree search.
\newblock In \emph{International Conference on Machine Learning}, pages
  1819--1827. PMLR, 2017.

\bibitem[Kovar{\'{\i}}k et~al.(2019)Kovar{\'{\i}}k, Schmid, Burch, Bowling, and
  Lis{\'{y}}]{fosg}
V.~Kovar{\'{\i}}k, M.~Schmid, N.~Burch, M.~Bowling, and V.~Lis{\'{y}}.
\newblock Rethinking formal models of partially observable multiagent decision
  making.
\newblock \emph{CoRR}, abs/1906.11110, 2019.
\newblock URL \url{http://arxiv.org/abs/1906.11110}.

\bibitem[Kova{\v{r}}{\'\i}k et~al.(2019)Kova{\v{r}}{\'\i}k, Schmid, Burch,
  Bowling, and Lis{\`y}]{kovavrik2019rethinking}
V.~Kova{\v{r}}{\'\i}k, M.~Schmid, N.~Burch, M.~Bowling, and V.~Lis{\`y}.
\newblock Rethinking formal models of partially observable multiagent decision
  making.
\newblock \emph{arXiv preprint arXiv:1906.11110}, 2019.

\bibitem[Lerer et~al.(2020)Lerer, Hu, Foerster, and Brown]{sparta}
A.~Lerer, H.~Hu, J.~N. Foerster, and N.~Brown.
\newblock Improving policies via search in cooperative partially observable
  games.
\newblock In \emph{AAAI}, 2020.

\bibitem[Morav{\v{c}}{\'\i}k et~al.(2017)Morav{\v{c}}{\'\i}k, Schmid, Burch,
  Lis{\`y}, Morrill, Bard, Davis, Waugh, Johanson, and
  Bowling]{moravvcik2017deepstack}
M.~Morav{\v{c}}{\'\i}k, M.~Schmid, N.~Burch, V.~Lis{\`y}, D.~Morrill, N.~Bard,
  T.~Davis, K.~Waugh, M.~Johanson, and M.~Bowling.
\newblock Deepstack: Expert-level artificial intelligence in heads-up no-limit
  poker.
\newblock \emph{Science}, 356\penalty0 (6337):\penalty0 508--513, 2017.

\bibitem[Nayyar et~al.(2014)Nayyar, Mahajan, and Teneketzis]{nayyar}
A.~Nayyar, A.~Mahajan, and D.~Teneketzis.
\newblock The {Common}-{Information} {Approach} to {Decentralized} {Stochastic}
  {Control}.
\newblock Springer Verlag, 2014.
\newblock \doi{10.1007/978-3-319-02150-8_4}.

\bibitem[Oliehoek(2013)]{oliehoek2013sufficient}
F.~A. Oliehoek.
\newblock Sufficient plan-time statistics for decentralized pomdps.
\newblock In \emph{Twenty-Third International Joint Conference on Artificial
  Intelligence}. Citeseer, 2013.

\bibitem[Rashid et~al.(2018)Rashid, Samvelyan, Schroeder, Farquhar, Foerster,
  and Whiteson]{rashid2018qmix}
T.~Rashid, M.~Samvelyan, C.~Schroeder, G.~Farquhar, J.~Foerster, and
  S.~Whiteson.
\newblock Qmix: Monotonic value function factorisation for deep multi-agent
  reinforcement learning.
\newblock In \emph{International Conference on Machine Learning}, pages
  4295--4304. PMLR, 2018.

\bibitem[Samvelyan et~al.(2019)Samvelyan, Rashid, de~Witt, Farquhar, Nardelli,
  Rudner, Hung, Torr, Foerster, and Whiteson]{samvelyan19smac}
M.~Samvelyan, T.~Rashid, C.~S. de~Witt, G.~Farquhar, N.~Nardelli, T.~G.~J.
  Rudner, C.-M. Hung, P.~H.~S. Torr, J.~Foerster, and S.~Whiteson.
\newblock {The} {StarCraft} {Multi}-{Agent} {Challenge}.
\newblock \emph{CoRR}, abs/1902.04043, 2019.

\bibitem[Silver et~al.(2016)Silver, Huang, Maddison, Guez, Sifre, Van
  Den~Driessche, Schrittwieser, Antonoglou, Panneershelvam, Lanctot,
  et~al.]{silver2016mastering}
D.~Silver, A.~Huang, C.~J. Maddison, A.~Guez, L.~Sifre, G.~Van Den~Driessche,
  J.~Schrittwieser, I.~Antonoglou, V.~Panneershelvam, M.~Lanctot, et~al.
\newblock Mastering the game of go with deep neural networks and tree search.
\newblock \emph{Nature}, 529\penalty0 (7587):\penalty0 484, 2016.

\bibitem[Silver et~al.(2017)Silver, Schrittwieser, Simonyan, Antonoglou, Huang,
  Guez, Hubert, Baker, Lai, Bolton, et~al.]{silver2017mastering}
D.~Silver, J.~Schrittwieser, K.~Simonyan, I.~Antonoglou, A.~Huang, A.~Guez,
  T.~Hubert, L.~Baker, M.~Lai, A.~Bolton, et~al.
\newblock Mastering the game of go without human knowledge.
\newblock \emph{Nature}, 550\penalty0 (7676):\penalty0 354, 2017.

\bibitem[Silver et~al.(2018)Silver, Hubert, Schrittwieser, Antonoglou, Lai,
  Guez, Lanctot, Sifre, Kumaran, Graepel, et~al.]{silver2018general}
D.~Silver, T.~Hubert, J.~Schrittwieser, I.~Antonoglou, M.~Lai, A.~Guez,
  M.~Lanctot, L.~Sifre, D.~Kumaran, T.~Graepel, et~al.
\newblock A general reinforcement learning algorithm that masters chess, shogi,
  and go through self-play.
\newblock \emph{Science}, 362\penalty0 (6419):\penalty0 1140--1144, 2018.

\bibitem[Sokota(2020)]{sokota_solving_2020}
S.~Sokota.
\newblock Solving {Common}-{Payoff} {Games} with {Approximate} {Policy}
  {Iteration}, 2020.
\newblock URL
  \url{https://era.library.ualberta.ca/items/17edd0aa-ec13-4cd4-8e30-a77b5d8c5ccc}.

\bibitem[Sokota et~al.(2021)Sokota, Lockhart, Timbers, Davoodi, D'Orazio,
  Burch, Schmid, Bowling, and Lanctot]{sokota2021solving}
S.~Sokota, E.~Lockhart, F.~Timbers, E.~Davoodi, R.~D'Orazio, N.~Burch,
  M.~Schmid, M.~Bowling, and M.~Lanctot.
\newblock Solving common-payoff games with approximate policy iteration.
\newblock In \emph{Thirty-Fifth {AAAI} Conference on Artificial Intelligence,
  {AAAI} 2021, Thirty-Third Conference on Innovative Applications of Artificial
  Intelligence, {IAAI} 2021, The Eleventh Symposium on Educational Advances in
  Artificial Intelligence, {EAAI} 2021, Virtual Event, February 2-9, 2021},
  pages 9695--9703. {AAAI} Press, 2021.
\newblock URL \url{https://ojs.aaai.org/index.php/AAAI/article/view/17166}.

\bibitem[Sokota et~al.(2022)Sokota, Hu, Wu, Kolter, Foerster, and Brown]{bft}
S.~Sokota, H.~Hu, D.~J. Wu, J.~Z. Kolter, J.~N. Foerster, and N.~Brown.
\newblock A fine-tuning approach to belief state modeling.
\newblock In \emph{International Conference on Learning Representations}, 2022.
\newblock URL \url{https://openreview.net/forum?id=ckZY7DGa7FQ}.

\bibitem[Strouse et~al.(2021)Strouse, McKee, Botvinick, Hughes, and
  Everett]{fict-coplay}
D.~Strouse, K.~R. McKee, M.~M. Botvinick, E.~Hughes, and R.~Everett.
\newblock Collaborating with humans without human data.
\newblock \emph{CoRR}, abs/2110.08176, 2021.
\newblock URL \url{https://arxiv.org/abs/2110.08176}.

\bibitem[Sunehag et~al.(2018)Sunehag, Lever, Gruslys, Czarnecki, Zambaldi,
  Jaderberg, Lanctot, Sonnerat, Leibo, Tuyls, and
  Graepel]{sunehag_value-decomposition_2018}
P.~Sunehag, G.~Lever, A.~Gruslys, W.~M. Czarnecki, V.~Zambaldi, M.~Jaderberg,
  M.~Lanctot, N.~Sonnerat, J.~Z. Leibo, K.~Tuyls, and T.~Graepel.
\newblock Value-{Decomposition} {Networks} {For} {Cooperative} {Multi}-{Agent}
  {Learning} {Based} {On} {Team} {Reward}.
\newblock In \emph{Proceedings of the 17th {International} {Conference} on
  {Autonomous} {Agents} and {MultiAgent} {Systems}}, {AAMAS} '18, pages
  2085--2087, Richland, SC, July 2018. International Foundation for Autonomous
  Agents and Multiagent Systems.

\bibitem[Tesauro(1995)]{tesauro1995temporal}
G.~Tesauro.
\newblock Temporal difference learning and td-gammon.
\newblock \emph{Communications of the ACM}, 38\penalty0 (3):\penalty0 58--68,
  1995.

\bibitem[Yu et~al.(2021)Yu, Velu, Vinitsky, Wang, Bayen, and
  Wu]{yu2021surprising}
C.~Yu, A.~Velu, E.~Vinitsky, Y.~Wang, A.~Bayen, and Y.~Wu.
\newblock The surprising effectiveness of mappo in cooperative, multi-agent
  games, 2021.

\bibitem[Zhang et~al.(2022)Zhang, Farina, and Sandholm]{2team0s}
B.~H. Zhang, G.~Farina, and T.~Sandholm.
\newblock Team belief {DAG} form: {A} concise representation for
  team-correlated game-theoretic decision making.
\newblock \emph{CoRR}, abs/2202.00789, 2022.
\newblock URL \url{https://arxiv.org/abs/2202.00789}.

\end{thebibliography}
